\documentclass[10pt]{article} 

\usepackage[preprint]{rlj}           

\usepackage{amssymb}            
\usepackage{mathtools}          
\usepackage{mathrsfs}           
\usepackage{graphicx}           
\usepackage{subcaption}         
\usepackage[space]{grffile}     
\usepackage{url}                
\usepackage{lipsum}             

\usepackage[utf8]{inputenc} 
\usepackage{booktabs}       
\usepackage{nicefrac}       
\usepackage{microtype}      
\usepackage{pdfpages}
\graphicspath{ {./res/} }
\usepackage{multirow}
\usepackage{dirtytalk}
\usepackage{float} 

\usepackage{siunitx}

\usepackage[capitalize,noabbrev]{cleveref}

\usepackage{overpic}

\title{Mapping representations in Reinforcement Learning via Semantic Alignment for Zero-Shot Stitching}

\setrunningtitle{Enter Your Running Title Here}

\author{Antonio Pio Ricciardi \textsuperscript{1}, Valentino Maiorca\textsuperscript{1}, Luca Moschella\textsuperscript{1}, Riccardo Marin\textsuperscript{2}, Emanuele Rodolà\textsuperscript{1}}

\emails{ricciardi@di.uniroma1.it}

\affiliations{
$^{1}$\textbf{Sapienza, Uniersity of Rome}\\
$^{2}$\textbf{University of Tübingen, Germany}\\
}

\contribution{
    Provide a succinct but precise list of the contribution(s) of the paper. Use contextual notes to avoid implications of contributions more significant than intended and to clarify and situate the contribution relative to prior work (see the examples below). If there is no additional context, enter ``None''. Try to keep each contribution to a single sentence, although multiple sentences are allowed when necessary. If using complete sentences, include punctuation. If using a single sentence fragment, you may omit the concluding period. A single contribution can be sufficient, and there is no limit on the number of contributions. Submissions will be judged mostly on the contributions claimed on their cover pages and the evidence provided to support them. Major contributions should not be claimed in the main text if they do not appear on the cover page. Overclaiming can lead to a submission being rejected, so it is important to have well-scoped contribution statements on the cover page.
    }
    {
    None
    }

\contribution{
    The submission template for submissions to RLJ/RLC 2025
    }
    {
    Built from previous RLC/RLJ, ICLR, and TMLR submission templates
    }

\contribution{
    \textit{{[}Example of one contribution and corresponding contextual note for the paper ``Policy gradient methods for reinforcement learning with function approximation'' \citep{Sutton2000}.]}\\ This paper presents an expression for the policy gradient when using function approximation to represent the action-value function.
    }
    {
    Prior work established expressions for the policy gradient without function approximation \citep{Williams1992}.
    }


\summary{The summary appears on the cover page. Although it can be identical to the abstract, it does not have to be. One might choose to omit the stated contributions in the Summary, given that they will be stated in the box below. The original abstract may also be extended to two paragraphs. The authors should ensure that the contents of the cover page fit entirely on a single page. The cover page does \textbf{not} count towards the 8--12 page limit.

\lipsum[1]
}

\newcommand{\AR}[1]{{\textcolor{red}{[\textbf{AR:} #1]}}}

\definecolor{cellblue}{RGB}{173,216,230}

\begin{document}

\maketitle  

\begin{abstract}
Deep Reinforcement Learning (RL) models often fail to generalize when even small changes occur in the environment’s observations or task requirements. Addressing these shifts typically requires costly retraining, limiting the reusability of learned policies. In this paper, we build on recent work in semantic alignment to propose a zero‐shot method for mapping between latent spaces across different agents trained on different visual and task variations. Specifically, we learn a transformation that maps embeddings from one agent’s encoder to another agent’s encoder without further fine‐tuning. Our approach relies on a small set of “anchor” observations that are semantically aligned, which we use to estimate an affine or orthogonal transform. Once the transformation is found, an existing controller trained for one domain can interpret embeddings from a different (existing) encoder in a zero‐shot fashion, skipping additional trainings. We empirically demonstrate that our framework preserves high performance under visual and task domain shifts. 
We empirically demonstrate zero-shot stitching performance on the CarRacing environment with changing background and task.
By allowing modular re‐assembly of existing policies, it paves the way for more robust, compositional RL in dynamically changing environments.    
\end{abstract}

\section{Introduction}

\begin{figure}[h]
    \centering
    \begin{overpic}[trim=0cm 0cm 0cm 0cm,clip,angle=0,origin=c,width=.4\linewidth]{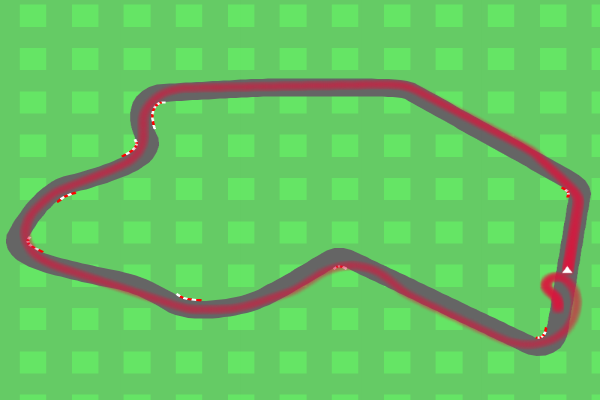}
        %
        \put(107, 32){$\mathbf{\to}$}
    \end{overpic}\hspace{1cm}
    \begin{overpic}[trim=0cm 0cm 0cm 0cm,clip,angle=0,origin=c,width=.4\linewidth]{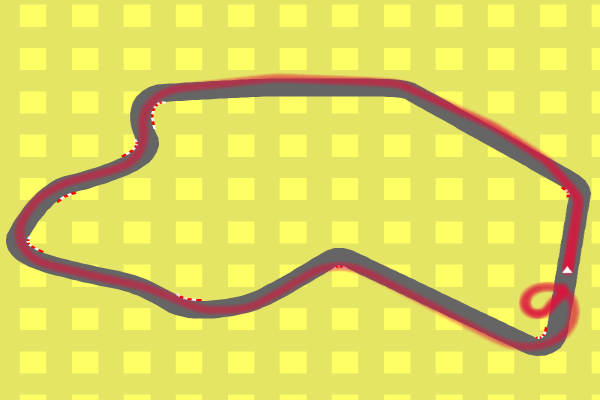}
        %
    \end{overpic}
    \caption{Using translation methods, a controller trained on an environment with a given visual variation \textit{(left)} can be reused without any training or fine-tuning on a different environment (\textit{right}) with comparable performance. In red we see the trajectory of a car driven by the same controller when connected to two different encoders, one for each visual variation.
    }
    \label{fig:teaser}
\end{figure}

Deep Reinforcement Learning (RL) has enabled agents to achieve remarkable performance in complex decision-making tasks, from robotic manipulation to high-dimensional games (Mnih et al., 2015; Silver et al., 2017). 
Although recent RL techniques achieved strong improvements over sample efficiency \citep{yarats2021drqv2, kostrikov2020image}, training new agents remains a costly process, both in computational and temporal terms.
Despite these advances, most methods still require at least partial retraining when dealing with domain shifts such as visual appearance, reward functions, or action spaces \citep{pmlr-v97-cobbe19a, zhang2020learning}. These domain changes typically require expensive retraining, which can be prohibitive for real-world settings that require millions of interactions.

A variety of approaches have been proposed to address these shifting conditions. Domain randomization \citep{tobin2017domain, sadeghi2016cad2rl} trains agents across diverse visual styles or physics settings, promoting invariant features but demanding broader coverage of possible variations. Multi-task RL \citep{parisotto2015actor, teh2017distral} attempts to learn shared representations across multiple tasks.

In the supervised setting, recent representation learning techniques \citep{Moschella2022-yf,maiorca2023latent, norelli2022b, cannistraci2023bricks}, show that it is possible to zero-shot recombine encoders and decoders to perform new tasks across different modalities (images, text..) and tasks (classification, reconstruction) and even architectures.
In RL, methods adopting the relative representation framework \citep{Moschella2022-yf} have shown promising results in adapting encoders to different controllers with zero or few-shots adaptation, for robotic control from proprioceptive states \citep{jian2021adversarial} or for playing games in the Gymnasium suite \citep{towers2024gymnasium} from pixels \citep{ricciardi2025r3lrelativerepresentationsreinforcement}.
These methods, however, still require training models to use the new relative representations.

By contrast, \cite{maiorca2023latent} suggest that modules from independently trained neural networks can be connected via a simple linear or affine transformation, with no training constraint or fine-tuning required, if such transformations can be reliably estimated from a small set of “anchor” samples, pairs of states or observations deemed semantically equivalent.

Our main contribution is the implementation of a RL method based on semantic alignment to map between latent spaces of different neural models, so that their encoders and controllers can be stitched with the goal of creating new agents that can act on visual-task combinations never seen together in training. This includes the use of the transformations to map modules from different networks, and the collection of anchor samples used to estimate these transformations. We call our method Semantic Alignment for Policy Stitching (\textbf{SAPS}).
We perform analyses and empirical tests on the CarRacing and LunarLander environments to show the performance of new agents created via zero-shot stitching of encoders and controllers trained on different visual-task variations, demonstrating significant gains compared to existing zero-shot methods.
\section{Related Work}

\paragraph{Domain Adaptation and Generalization in RL}

One line of work tackling the mismatch between training and deployment environments focuses on domain adaptation. Early solutions often rely on domain randomization, which exposes agents to a broad range of visual or dynamical variations during training so that they learn features robust to such changes \citep{tobin2017domain, sadeghi2016cad2rl}. Similarly, data augmentation techniques \citep{yarats2021drqv2, laskin2020reinforcement} modify raw input frames (e.g., random crops, color jitter) to improve out-of-distribution performance \citep{kostrikov2020image, yarats2021drqv2}.
However, these methods either demand extensive coverage of possible variations or can be extremely resource intensive.

\paragraph{Multi-task and Modular RL}
A common strategy to reuse knowledge across related tasks is multi-task or meta-RL, where a single agent is trained on multiple environments \citep{teh2017distral, finn2017model}. While this can yield more robust representations, it often demands joint training on all tasks, which might be impractical when new tasks appear over time. Transfer RL \citep{taylor2009transfer} instead aims to expedite learning in a target task by leveraging agents trained on a source task, reusing features or parameters \citep{barreto2017successor, killian2017robust}, value functions (\cite{tirinzoni2018transfer, liu2021learning}, sub-policies \citep{fernandez2006probabilistic, devin2017learning}.
Other approaches exploit networks modularity, by designing agents as composition of planner and actuator modules \citep{karkus2020beyond} or by combining submodules to solve harder tasks \citep{mendez2022modular, russell2003q, simpkins2019composable}. These, however, require defining ad-hoc architectures to work. Our method instead focuses on zero-shot reuse of already existing models, while also eliminating the need for environment interaction or fine-tuning to adapt an already-trained module to a new domain.

\paragraph{Representation Learning}
Exploiting representation learning techniques is another approach to mitigate the complexity in RL. Some approaches propose to isolate visual or observational factors from task-dependent decision-making \citep{oord2018representation}, or learning a robust policy given a source domain \citep{higgins2017darla}. Other ignore task-irrelevant features via invariant encoders, using bisimulation metrics as training constraints \citep{zhang2020learning}, using supervised learning o train inverse dynamics models \citep{hansen2020self} or finetuning to match prior latent distributions \citep{yoneda2021invariance}.
However, these approaches focus exclusively on visual variations, leaving shifts in the underlying task unaddressed. While such fine-tuning can be more efficient than learning from scratch, it still requires re-optimization whenever the environment changes. In contrast, our proposed approach does not need re-training or fine-tuning. Instead, it only needs to encode small transformation between existing latent spaces.

\paragraph{Model Stitching}
Model stitching refers to the process of combining separate neural modules (e.g. encoders, decoders, or intermediate layers from independently trained models) to create a new, fully functional model. Initially studied in supervised learning as means to measure latent space similarity across different models, \citep{Lenc2014-gy, Bansal2021-oj, Csiszarik2021-yi}, models stitching recently is being used for zero-shot model reuse \citep{Moschella2022-yf, norelli2022b, maiorca2023latent, cannistraci2023bricks} by using sets of parallel data called \textit{anchors}, which establish a semantic correspondence between models. These are used in the relative representation \citep{Moschella2022-yf} framework to project latent spaces to a common space, and the semantic alignment framework \citep{maiorca2023latent} to directly estimate a mapping between latent spaces of different models.

Recently, the relative representation framework has been used to perform stitching between encoders and controllers in the context of vision-based imitation learning \citep{jian2024perception}, while in RL it has been used to perform few-shot and zero-shot stitching in robotic control from low dimensional proprioceptive states \citep{jian2023policy} or control from pixels in the Gymnasium suite \citep{ricciardi2025r3lrelativerepresentationsreinforcement}.
Although powerful, these methods require training or fine-tuning models or the decoder, so that it learns to act given \say{relative} inputs, while also requiring additional training constraints to not degrade end-to-end performance.
By contrast, our approach applies the idea of creating a mapping between layers by using anchor samples \citep{maiorca2023latent} and is directly applicable to already trained models, learning a transformation that maps between modules of different policies without further constraints or fine-tuning.

\newcommand{\enc}[0]{\phi}
\newcommand{\con}[0]{\psi}
\newcommand{\encmap}[2]{\mathcal{O}_{#1}^{#2} \mapsto \mathcal{X}_{#1}^{#2}}
\newcommand{\conmap}[3]{\mathcal{X}_{#1}^{#2} \mapsto \mathcal{A}_{#3}}
\newcommand{\ours}[0]{Trasl. \textbf{(Ours)}}

\section{Preliminaries}

We assume the underlying environment is a Markov decision process $\mathcal{M} = (\mathcal{S, A}, \mathcal{O}, R, P, \gamma)$, with state space $\mathcal{S}$, action space $\mathcal{A}$, input observations $o \in \mathcal{O}$ and the transition
function $P : \mathcal{S} \times \mathcal{A} \mapsto \mathcal{S}$ that defines a probability distribution over the next state given the current state and action. The function $\mathcal{R} : \mathcal{S} \times \mathcal{A} \mapsto \mathcal{R}$ assigns rewards, and $\gamma$ is the discount factor that reduces the importance of delayed rewards. The agent’s behavior is dictated by a policy $\pi : \mathcal{O} \rightarrow \mathcal{A}$, which receives an observation and selects an action at each state, and is trained to maximize the discounted returns
$\mathbb{E}\Bigl[\sum_{i=0}^{\infty} \gamma^{i} \mathcal{R}(\mathbf{s}_{i}, \mathbf{a}_{i})\Bigr]$.

\subsection{Background}
We are interested in scenarios where both training setups and agent behaviors can vary. We find it convenient to use the same notation introduced in \textsc{R3L} (Relative Representations for Reinforcement Learning) \citep{ricciardi2025r3lrelativerepresentationsreinforcement} to formalize these variations.

\paragraph{Environment variations}
We denote an environment by $\mathcal{M}{u}^{i} = (\mathcal{O}_{u}, T_{i})$. Here, $\mathcal{O}_{u}$ is the distribution of observations $o_{u}$, and $T_{i} : \mathcal{S}{i} \times \mathcal{A}_{i} \times \mathcal{R}_{i} \times P{i} \mapsto \mathcal{R}_{i}$  specifies the task. Two environments can differ in the distribution of observations (e.g., background color, camera perspective) or in the task itself (e.g., transition dynamics, action spaces, reward definitions).
Since agents must discover the task solely through reward feedback, any shift—whether in observations or task—can significantly alter their learned representations.

\paragraph{Agents}
Following \textsc{R3L}, each policy $\pi_{u}^{i}$ is typically obtained by end-to-end training on $\mathcal{M}{u}^{i}$. However, we emphasize a modular view of this policy:
\begin{align}
    \pi_{u}^{i}(o_{u}) \;=\; \con_{u}^{i}\bigl[\enc_{u}^{i}(o_{u})\bigr] \;=\; \con_{u}^{i}\bigl(\mathbf{x}_{u}^{i}\bigr)
\end{align}
where $\enc_{u}^{i} : \mathcal{O}_{u} \mapsto \mathcal{X}_{u}^{i}$ serves as the \textit{encoder} that processes raw observations (e.g., images), and $\con_{u}^{i} : \mathcal{X}_{u}^{i} \mapsto \mathcal{A}_{i}$ is the \textit{controller} that outputs actions based on the latent embedding $\mathbf{x}{u}^{i}$. This factorization disentangles observation-specific features (in $\enc_{u}^{i}$) from task-specific decision rules (in $\con_{u}^{i}$).

\paragraph{Latent Representation}
Now consider a second environment $\mathcal{M}_{v}^{j} = (\mathcal{O}_{v}, T_{j})$, where $\mathcal{O}_{v}$ differs from $\mathcal{O}_{u}$ only in visual style (e.g., a shifted color scheme), and suppose we have a policy $\pi_{v}^{j}$ trained on that environment. For two semantically corresponding observations $o_{u} \in \mathcal{O}_{u}$ and $o_{v} \in \mathcal{O}_{v}$, the respective latent embeddings differ:
\begin{align}
\enc_{u}^{i}(o_{u}) \; \neq \; \enc_{v}^{j}(o_{v}) \quad \Longrightarrow \quad \mathbf{x}_{u}^{i} \;\neq\; \mathbf{x}_{v}^{j}
\end{align}
In the next section, we describe how to map one latent space onto another to enable zero-shot stitching of encoders and controllers trained in different visual and task domains, without additional training. 

\section{SAPS: Semantic Alignment for Policy Stitching}\label{sec:method-alignment}
Relative representations \citep{Moschella2022-yf}, used as a base for zero-shot stitching in R3L, involve computing a distance function between a set of samples, called \say{anchors}, to project the output of each encoder to a shared latent space, enabling the subsequent training of a universal policy. Semantic alignment, instead, estimates a direct mapping between latent spaces.

Consider the environment $\mathcal{M}_u^j$ for which no dedicated policy exists. However, we do have an encoder $\phi_u^i$ and a controller $\psi_v^j$, extracted from policies $\pi_u^i$ and $\pi_v^j$, respectively. 
We estimate an affine transformation $\tau_u^v$: $\mathcal{X}_{u}^{i} \mapsto \mathcal{X}_{v}^{j}$, mapping embeddings produced by $\phi_u^i$ into the space of $\pi_v^j$. This yields a new latent space:

\begin{align}
    & \tau_u^v(\enc_u^i(\mathbf{o}_u)) \approx \enc_v^j(\mathbf{o}_v)\\
    & \tau_u^v(\mathbf{x}_{u}^i) \approx \mathbf{x}_{v}^j
\end{align}

that is compatible with the existing $\psi_v^j$.
This enables the stitching of encoders and controllers from $\pi_u^i$ and $\pi_v^j$, respectively, to obtain a new policy $\tilde{\pi}_u^j$ that can act in $\mathcal{M}_u^j$, without additional training:
\begin{equation}\label{eq:2}
    \tilde{\pi}_u^j(o_u) = \con_v^j[\tau_u^v(\enc_u^i(\mathbf{o}_u))]
\end{equation}

\paragraph{Estimating $\tau$}
As in \cite{maiorca2023latent}, assume to be given latent spaces $\mathbf{X}_u$ and $\mathbf{X}_v$ which here correspond to the embedding of two visual variations in the space of observations.
We use SVD to obtain an affine transformation $\tau_u^v(\mathbf{x}_u) = \mathbf{R} \mathbf{X}_u + \mathbf{b}$.

\paragraph{Collecting the Dataset.}
The anchor embeddings $\mathbf{X}_u$ and $\mathbf{X}_v$ derive from sets of anchor points $\mathbf{A}_u$ and $\mathbf{A}_v$. Following previous works \citep{maiorca2023latent, Moschella2022-yf, ricciardi2025r3lrelativerepresentationsreinforcement} anchor pair ($\mathbf{a}_u$, $\mathbf{a}_v$) must share a semantic correspondence, meaning both samples represent the same underlying concept (e.g., the same spatial position in a racing track, viewed under two different visual styles).
In supervised learning contexts, anchor pairs can come from paired datasets (e.g., bilingual corpora). In the context of online RL, however, such datasets do not naturally exist. Hence, we collect datasets sharing a correspondence.
This correspondence can be obtained by either rolling out a policy and replaying the same set of actions with different visual variations, as already done in \cite{jian2023policy, ricciardi2025r3lrelativerepresentationsreinforcement}, or by simply applying visual transformations to the image in pixel space. This yields corresponding observation sets $\mathbf{A}_u$ and $\mathbf{A}_v$ that can be embedded by each domain’s encoder to create $\mathbf{X}_u$ and $\mathbf{X}_v$. Finally, we solve for $\tau_u^v$ using the SVD-based procedure above.


In our context, we assume that an agent trained end-to-end to solve a specific task in a specific environment will generate a comprehensive set of observations, providing a reasonable approximation of the entire latent space. Nevertheless, forcing the agent to explore more could be beneficial in this context.
In our experiments, we gather parallel samples either by directly translating the observation in pixel space, when there is a well-defined known visual variation between the environments, or by replaying the same sequence of actions in both environments, that in this case must be deterministic and initialized with the same random seed. We leave to future research other possible approximation techniques for translating observations between different environments.
\section{Experiments}\label{sec:experiments}
We now evaluate SAPS using both qualitative and quantitative analyses. We first compare its zero-shot performance to R3L on benchmark tasks, then 
an analysis of how our alignment approach behaves under different conditions.

\paragraph{Environments}
Our agents act by receiving pixel images as input observation, consisting of four consecutive $84 \times 84$ RGB images, stacked along the channel dimension to capture dynamic information such as velocity and acceleration.
We consider environments where we can freely change visual features (background color, camera perspective) or task (rewards, dynamics), therefore we use CarRacing \citep{klimov2016carracing} and LunarLander as both implemented in R3L.
CarRacing requires the agent to drive in a track using pixel observations, whose variations can be in the background color or the target speed, while LunarLander requires the agent to land on a platform, with variations comprising background color and different gravities.
No context is provided, hence the agents do not receive any information about the task.

\paragraph{Baselines}
We mainly compare SAPS to (R3L), another zero-shot stitching method using relative representations whose approach is similar to ours.
For an additional baseline we also compare to naive zero-shot stitching, where we stitch encoders and controllers with no additional processing, to showcase the progress reached by the methods performing latent alignment techniques.

\subsection{Zero-shot stitching comparison}\label{sec:zero-shot-stitching}
We follow the empirical analysis performed in R3L. We define the encoder as the network up to the first flatten layer after the convolutional blocks, with the remaining layers constituting the controller. The zero-shot stitching evaluation is conducted on visual-task variations that were not seen together during training. Encoders and controllers must match to the specific conditions they were trained on. For example, an encoder trained with a green background should be used in such an environment, and a controller developed for low-speed driving should be applied to that task.

\paragraph{Stitching table}
Encoders and controllers from different policies trained under different conditions, can be independently assembled through zero-shot stitching. 
In \Cref{sec:method-alignment} we defined how to estimate and use a transformation to then perform the stitching, defining the models as formed by encoders and controllers. 

We present zero-shot stitching results in \Cref{tab:carracing-stitching_performance} and \cref{tab:lunarlander-stitching_performance}. 
The testing procedure to perform stitching is as follows:
Given models trained with various seeds and different visual-task variations, we decompose encoders and controllers of a model, then connect each encoder with controllers that were trained either on different visual variations or different seeds, using the transformation to ensure compatibility between layers.
Each cell reports the average score obtained across different stitched seeds.
Therefore, if we have four visual variations, six task variations and five seeds, for each cell we perform $6 \times 5 = 30$ stitching tests, for a total of 120 across the entire table for a single stitching method. Each combination is tested over 10 different tracks.

In CarRacing, SAPS consistently achieves performance comparable to end-to-end scores across all tested visual and task variations, even in the challenging slow scenario where R3L experiences a noticeable drop. In LunarLander there is a considerable drop in the mean average performance, which is mainly due to some models highly underperforming, decreasing the mean value. This is, however, a strong improvement over R3L, for which we were not even able to train models, strongly highlighting the advantages of being able to perform zero-shot stitching from already trained, standard models.
This table underscores SAPS’s robustness and adaptability in assembling agents for new environment variations, surpassing existing methods.

SAPS instead performs much worse in LunarLander with respect to end-to-end agents, although improving upon the naive stitching approach. We think that this might reflect how sensitive LunarLander is to even small latent-space mismatches: a slight misalignment in the latent space could cause the landing procedure to strongly deviate in its trajectory, causing crashes and large penalties. Here, R3L scores are missing because we were not able to train models using this method.

\begin{table}[t!]
    \caption{Stitching performance in \textbf{CarRacing}, comparing \textbf{SAPS (ours)} to other methods, averaged over 5 seeds, with standard deviations.}
    \label{tab:carracing-stitching_performance}
    \resizebox{\textwidth}{!}{
    \begin{tabular}{ccccccccccc}
    \toprule
    & & & \multicolumn{6}{c}{\textbf{Controller}} & \\
    \cmidrule{4-9}
    & & & \multicolumn{3}{c}{\textbf{Visual Variations} (task standard)} & \multicolumn{3}{c}{\textbf{Task Variations} (green)} \\
    \cmidrule(r){4-6} \cmidrule(l){7-9}
    & & & \texttt{green} & \texttt{red} & \texttt{blue} &  \texttt{slow} & \texttt{scrambled} & \texttt{no idle} \\
    \cmidrule(r){1-6} \cmidrule(l){7-9}
    \multirow{12}{*}{\rotatebox{90}{\textbf{Encoder}}} 
    & \multirow{3}{*}{\rotatebox{90}{\texttt{green}}} 
    & \emph{Naive} & $175 \pm 304$ & $167 \pm 226$ & $-4 \pm 79$ &  $148 \pm 328$ & $106 \pm 217$ & $213 \pm 201$ \\
    & & \emph{R3L} & $781 \pm 108$ & $787 \pm 62$ & $794 \pm 61$ & $268 \pm 14$ & $781 \pm 126$ & $824 \pm 82$ \\
    & & \textbf{\emph{SAPS} (ours)} & $\mathbf{822 \pm 62}$ & $\mathbf{786 \pm 82}$ & $\mathbf{829 \pm 49}$ & $\mathbf{764 \pm 287}$ & $\mathbf{846 \pm 66}$ & $\mathbf{781 \pm 72}$ \\[1.5ex]
    & \multirow{3}{*}{\rotatebox{90}{\texttt{red}}} 
    & \emph{Naive} & $157 \pm 248$ & $43 \pm 205$ & $22 \pm 112$  & $83 \pm 191$ & $138 \pm 244$ & $252 \pm 228$ \\
    & & \emph{R3L} & $810 \pm 52$ & $776 \pm 92$ & $803 \pm 58$ & $476 \pm 430$ & $790 \pm 72$ & $817 \pm 69$ \\
    & & \textbf{\emph{SAPS} (ours)} & $\mathbf{859 \pm 41}$ & $\mathbf{807 \pm 52}$ & $\mathbf{809 \pm 60}$ & $\mathbf{824 \pm 192}$ & $\mathbf{838 \pm 52}$ & $\mathbf{853 \pm 50}$ \\[1.5ex]
    & \multirow{3}{*}{\rotatebox{90}{\texttt{blue}}} 
    & \emph{Naive} & $137 \pm 225$ & $130 \pm 274$ & $11 \pm 122$ & $95 \pm 128$ & $138 \pm 224$ & $144 \pm 206$ \\
    & & \emph{R3L} & $791 \pm 64$ & $793 \pm 40$ & $792 \pm 48$  & $564 \pm 440$ & $804 \pm 41$ & $828 \pm 50$ \\
    & & \textbf{\emph{SAPS} (ours)} & $\mathbf{839 \pm 57}$ & $\mathbf{808 \pm 70}$ & $\mathbf{814 \pm 52}$  & $\mathbf{746 \pm 319}$ & $\mathbf{832 \pm 60}$ & $\mathbf{808 \pm 62}$ \\[1.5ex]
    & \multirow{3}{*}{\rotatebox{90}{\texttt{far}}} 
    & \emph{Naive} & $152 \pm 204$ & $65 \pm 180$ & $2 \pm 152$ & $-49 \pm 9$ & $351 \pm 97$ & $349 \pm 66$ \\
    & & \emph{R3L} & $527 \pm 142$ & $605 \pm 118$ & $592 \pm 86$ & $303 \pm 100$ & $594 \pm 39$ & $673 \pm 91$ \\
    & & \textbf{\emph{SAPS} (ours)} & $\mathbf{714 \pm 45}$ & $\mathbf{712 \pm 71}$ & $\mathbf{727 \pm 52}$ & $\mathbf{762 \pm 131}$ & $\mathbf{738 \pm 44}$ & $\mathbf{626 \pm 77}$ \\
    \bottomrule
    \end{tabular}
    }
\end{table}

\begin{table}[t!]
    \centering
    \caption{Stitching performance comparing \textbf{SAPS (ours)} to other methods, in the \textbf{LunarLander} environment. Scores are averaged over 5 seeds, with standard deviations. For comparison, end-to-end models achieve $221 \pm 86$ for the white background, $192 \pm 30$ for the red background with gravity -10; for the white background, for gravity -3. R3L results are absent because we were not able to train models for it.}
    \label{tab:lunarlander-stitching_performance}
    \resizebox{0.6\textwidth}{!}{
    \begin{tabular}{ccccccc}  
    \toprule
    & \multicolumn{3}{c}{} & \multicolumn{3}{c}{\textbf{Controller}} \\
    \cmidrule(l){5-7}
    & \multicolumn{3}{c}{} & \multicolumn{2}{c}{\textbf{Gravity: -10}} & \textbf{Gravity: -3} \\
    \cmidrule(r){5-6} \cmidrule(l){6-7}
    & & & & \texttt{white} & \texttt{red} & \texttt{white} \\
    \midrule
    \multirow{5}{*}{\rotatebox{90}{\textbf{Enc}}} & \multirow{5}{*}{\rotatebox{90}{\textbf{Gravity -10}}}
    & \multirow{3}{*}{\rotatebox{90}{\texttt{white}}} 
    & \emph{Naive} & $-413 \pm 72$ & $-390 \pm 176$ & $-276 \pm 8$ \\
    & & & \textbf{\emph{SAPS} (ours)} & $\mathbf{19 \pm 56}$ & $\mathbf{8 \pm 60}$ & $-242 \pm 51$ \\[1.5ex]
    & & \multirow{3}{*}{\rotatebox{90}{\texttt{red}}} 
    & \emph{Naive} & $-444 \pm 116$ & $-403 \pm 109$ & $-271 \pm 18$ \\
    & & & \textbf{\emph{SAPS} (ours)} & $\mathbf{52 \pm 44}$ & $\mathbf{33 \pm 61}$ & $-204 \pm 71$ \\
    \bottomrule
    \end{tabular}
    }
\end{table}

\subsection{Latent Space Analysis}\label{sec:latent-analysis}

To assess how effectively our proposed method (SAPS) aligns latent representations across different visual or task variations, we perform both qualitative and quantitative evaluations.

\begin{figure}[t!]
    \centering
    \includegraphics[width=\linewidth]{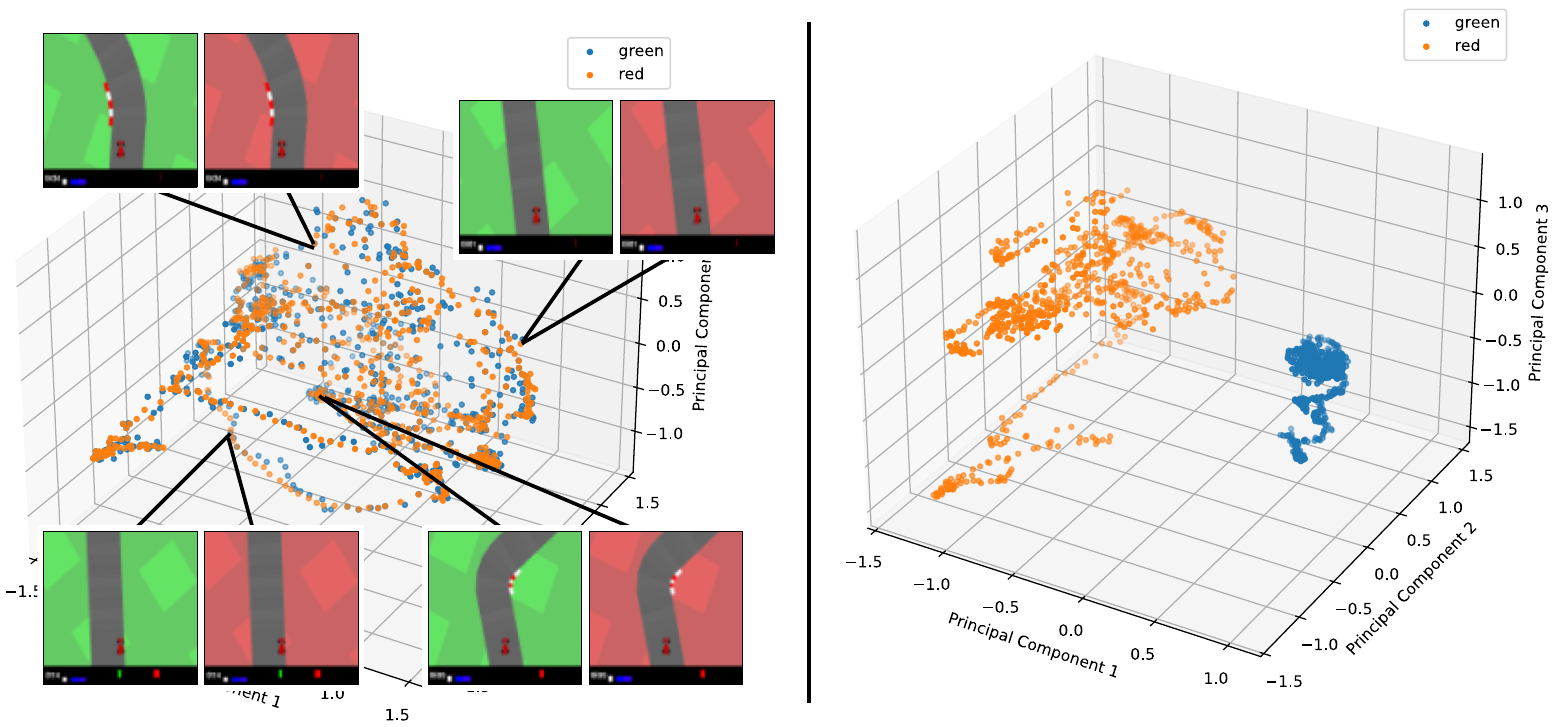}
    \caption{PCA visualization of encoder outputs. On the left, we illustrate how an affine alignment can effectively map one latent space to another: same frames with different backgrounds (green/red) cluster together, as indicated by the embedded screenshots. On the right, the source, unaligned embeddings remain separated, highlighting the benefit of our alignment approach in unifying observations from different environment variations.}
    \label{fig:pca-carracing}
\end{figure}

\paragraph{Qualitative Visualization.}
Figure \ref{fig:pca-carracing} (left) presents a 3D PCA projection of latent embeddings for two CarRacing variations: green and red backgrounds. After applying an affine transformation learned by SAPS to the encoder outputs, the points corresponding to green and red observations become thoroughly intermixed in the shared latent space. This intermixing indicates that frames depicting the “same portion” of the track with different background colors now map to similar embeddings (see insets). Conversely, Figure \ref{fig:pca-carracing} (right) shows the unaligned embeddings, where green and red remain clearly separated. These results suggest that SAPS successfully bridges the gap between encoders, producing a unified representation space under visual variation.

\begin{figure}[t!]
    \centering
    \begin{subfigure}[b]{0.33\linewidth}
        \centering
        \includegraphics[width=\linewidth]{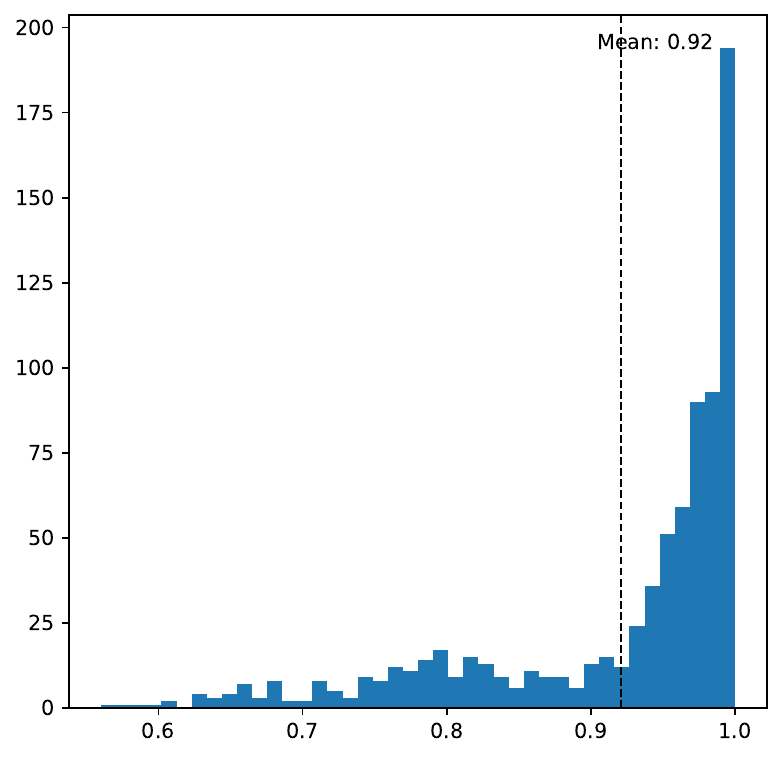}
        \caption{SAPS}
        \label{fig:frames-sim}
    \end{subfigure}%
    \begin{subfigure}[b]{0.33\linewidth}
        \centering
        \includegraphics[width=\linewidth]{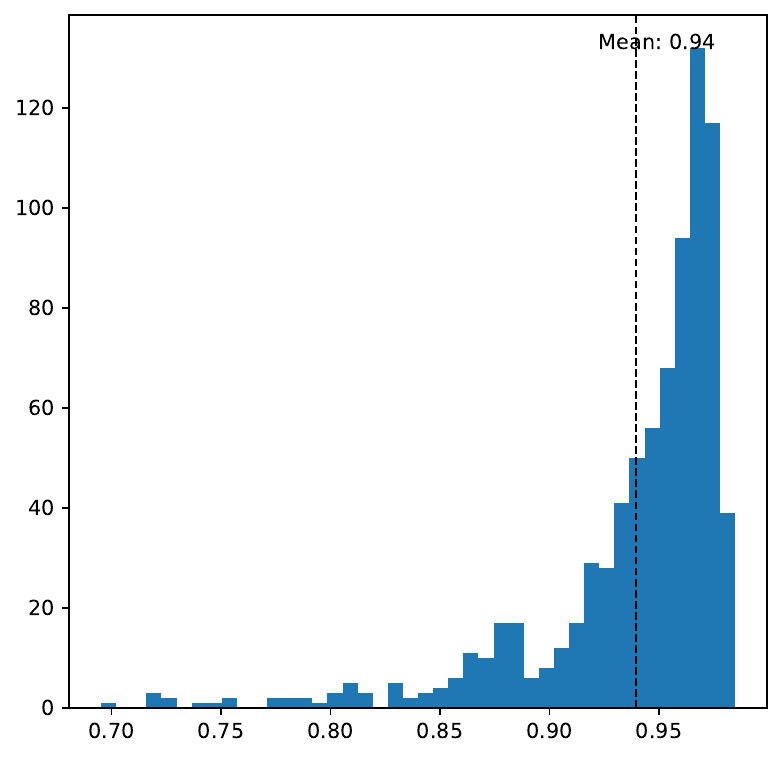}
        \caption{R3L}
        \label{fig:frames-sim}
    \end{subfigure}
    \begin{subfigure}[b]{0.33\linewidth}
        \centering
        \includegraphics[width=\linewidth]{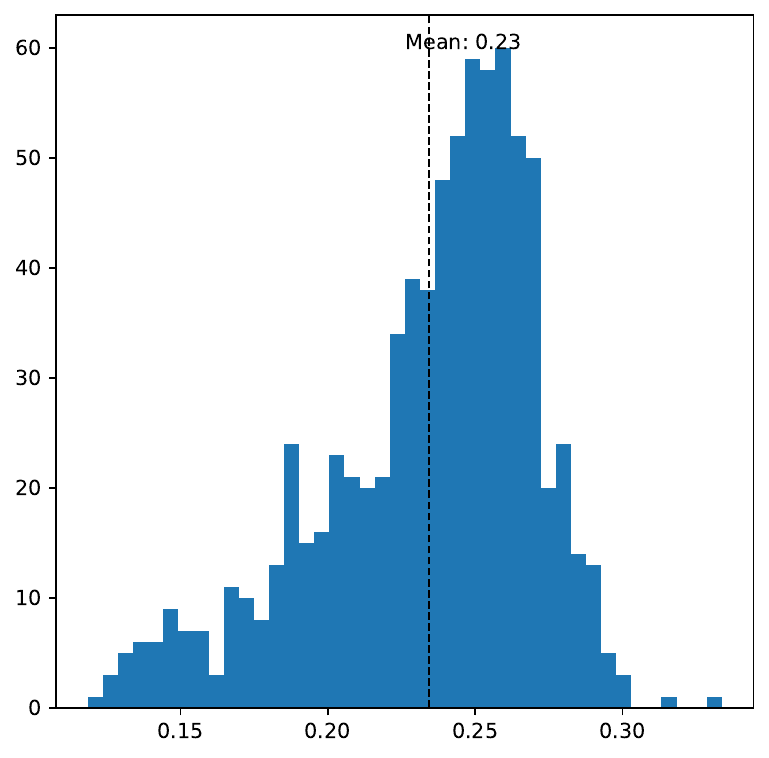}
        \caption{Naive}
        \label{fig:frames-sim}
    \end{subfigure}
        \begin{subfigure}[b]{0.33\linewidth}
        \centering
        \includegraphics[width=\linewidth]{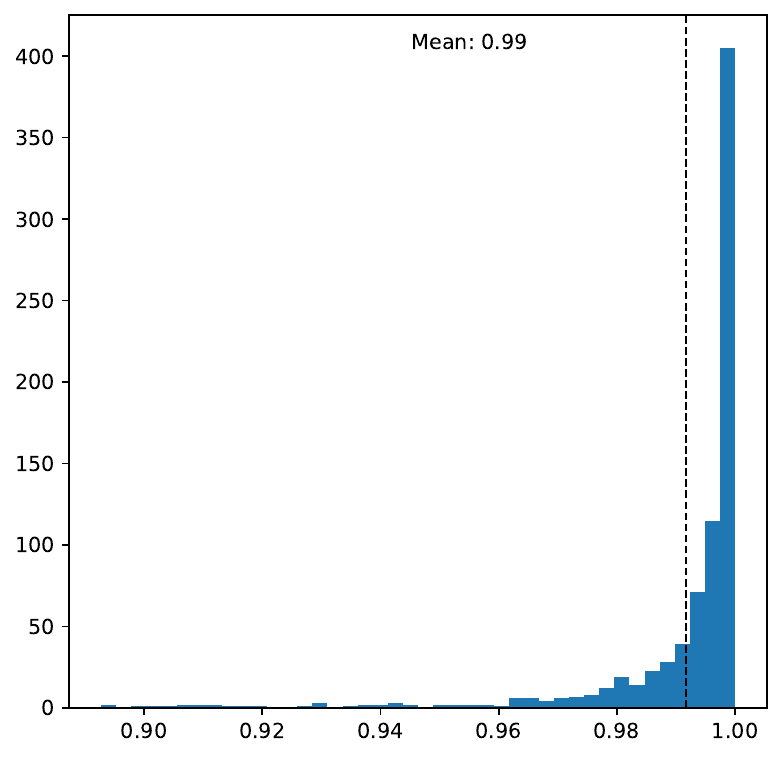}
        \caption{SAPS}
        \label{fig:frames-sim}
    \end{subfigure}%
    \begin{subfigure}[b]{0.33\linewidth}
        \centering
        \includegraphics[width=\linewidth]{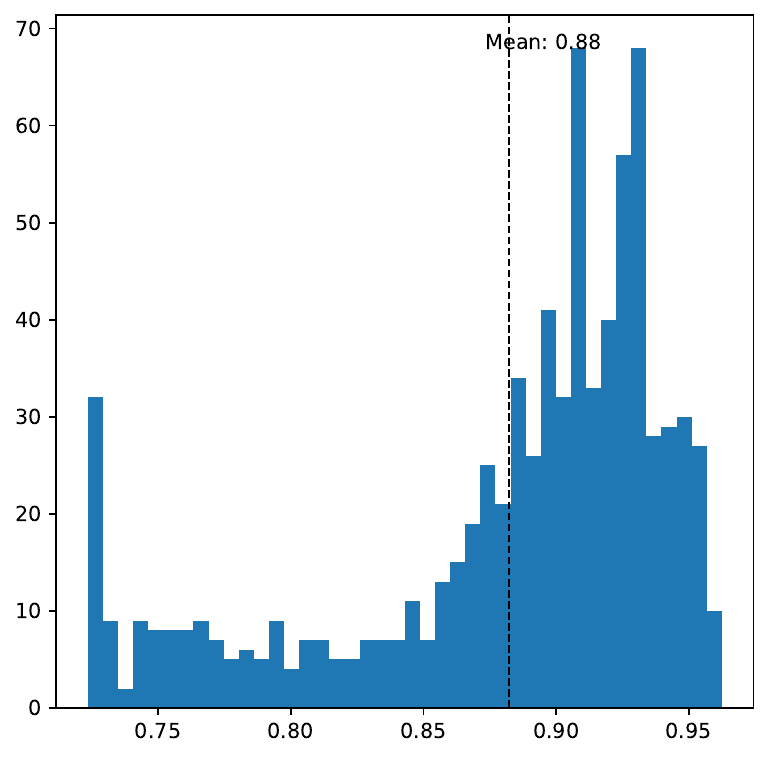}
        \caption{R3L}
        \label{fig:frames-sim}
    \end{subfigure}
    \begin{subfigure}[b]{0.33\linewidth}
        \centering
        \includegraphics[width=\linewidth]{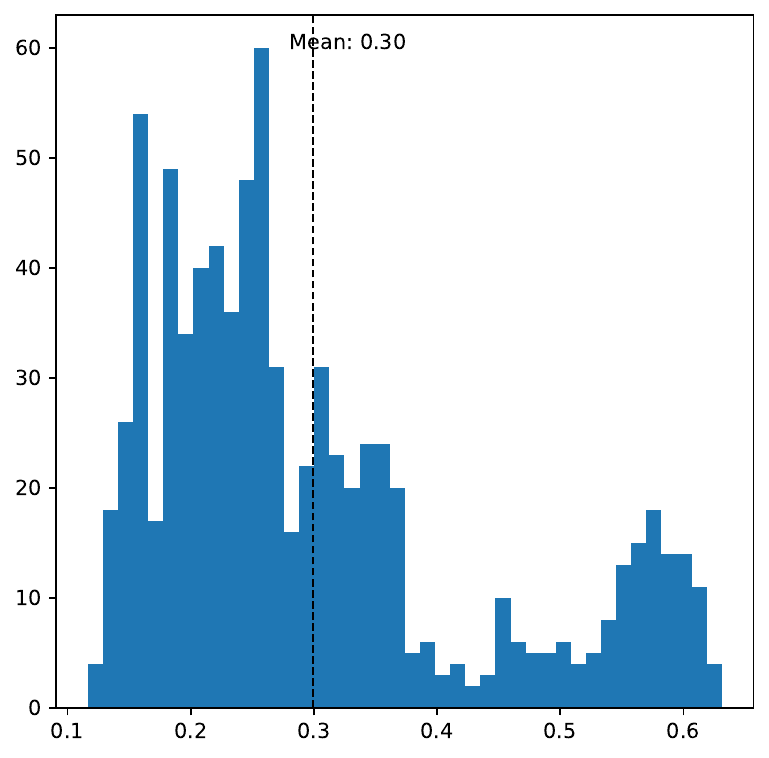}
        \caption{Naive}
        \label{fig:frames-sim}
    \end{subfigure}
    \caption{Histogram of pairwise cosine similarities between matched states from two different environment variations, for CarRacing (\textbf{top}) and LunarLander (\textbf{bottom}). Both SAPS and R3L show very high mean similarity along paired frames, indicating that corresponding observations in each variation map to nearly identical vectors. Mean similarity for encoders without any alignment or relative encoding is very low, emphasizing the utility of latent communication methods.}
    \label{fig:pairwise-histograms}
\end{figure}

\paragraph{Quantitative Similarities.}
In Figure~\ref{fig:pairwise-histograms}, we plot histograms of pairwise cosine similarity for matched frames from two different variations. Again, we compare \textbf{(a)~SAPS}, \textbf{(b)~R3L} and \textbf{(c)~Naive} stitching.
The top row shows CarRacing, while the bottom row is LunarLander. In both environments, SAPS and R3L achieve a much higher mean cosine similarity (e.g., 0.92–0.99) than the Naive baseline (0.23–0.30). This confirms that independently trained models can exhibit near-identical encodings for semantically identical frames once aligned, while “naive” combinations of encoders and controllers remain incompatible.

\paragraph{Discussion.}
Overall, these findings indicate that: (i) SAPS’ affine transformation effectively repositions points in the latent space, causing corresponding frames to map to nearly the same vector (Figures \ref{fig:pca-carracing}–\ref{fig:pairwise-histograms}); (ii) Compared to “Naive” reusability (no alignment) or purely relative approaches (R3L), SAPS achieves equivalent or better alignment without retraining models on a specialized representation format; (iii) The high average cosine similarity under SAPS confirms that visual variations (and, by extension, moderate task changes) can be handled by learning a lightweight transform from one latent space to another.

Hence, our latent space analysis demonstrates that SAPS successfully stitches together components from different RL models to produce a cohesive, unified representation, paving the way for zero-shot policy reuse in previously unseen environment variations.

\section{Conclusion and Future Directions}
\paragraph{Conclusion} We presented Semantic Alignment for Policy Stitching (SAPS), a simple yet effective method for zero-shot reuse of RL agents trained in different environments. By estimating a lightweight transformation that maps one encoder's latent space into another's, we can seamlessly “stitch” encoders and controllers, enabling new policies to handle unseen combinations of visual and task variations without retraining. Our experiments on CarRacing and LunarLander show SAPS achieves near end-to-end performance under diverse domain shifts, outperforming naive baselines and matching or exceeding more specialized zero-shot approaches (e.g., R3L) in many settings. This highlights the potential of direct latent-space alignment for compositional and robust RL.

\paragraph{Limitations and Future Works.}
Although SAPS effectively aligns latent representations across moderate environment variations, several open challenges remain. First, the method's reliance on affine transformations can falter in domains exhibiting larger gaps (for instance, tasks that differ drastically in reward structure or observation type) where a simple linear map may be insufficient. Second, the approach depends on anchors, which must be collected from both source and target environments; in highly stochastic domains, obtaining robust correspondences can be nontrivial. Third, the method was tested on relatively constrained settings, leaving it unclear how well it scales to real-world robotics or continuous domains with high state space complexity. Possible extensions include (i) automating or relaxing the anchor-collection procedure to handle more diverse or partially observable environments, and (ii) validating SAPS on robotics tasks with real sensors and complex dynamics, where retraining from scratch is particularly time-consuming. 

\bibliography{bibliography}
\bibliographystyle{rlj}

\end{document}